\documentclass[letterpaper]{article} 
\usepackage{aaai25}  
\usepackage{times}  
\usepackage{helvet}  
\usepackage{courier}  
\usepackage[hyphens]{url}  
\usepackage{graphicx} 
\usepackage{natbib}  
\usepackage{caption} 
\usepackage{booktabs}
\usepackage{multirow}
\usepackage{amssymb}
\usepackage{bbding}
\usepackage{algorithm}
\usepackage{algorithmic}
\usepackage{amsmath}
\usepackage{subfigure}
\usepackage{newfloat}
\usepackage{listings}
\usepackage{arydshln}
\DeclareCaptionStyle{ruled}{labelfont=normalfont,labelsep=colon,strut=off} 
\floatstyle{ruled}
\newfloat{listing}{tb}{lst}{}
\floatname{listing}{Listing}
\usepackage{xcolor}

 \usepackage{amsmath}
\title{DECT: Harnessing LLM-assisted Fine-Grained Linguistic Knowledge and Label-Switched and Label-Preserved Data Generation for Diagnosis of Alzheimer’s Disease}
\author{
    Tingyu Mo\textsuperscript{\rm 1},
    Jacqueline C. K. Lam\textsuperscript{\rm 1}\thanks{Corresponding Authors},
    Victor O.K. Li\textsuperscript{\rm 1}\footnotemark[1],
    Lawrence Y. L. Cheung\textsuperscript{\rm 2}
}
\affiliations{
    \textsuperscript{\rm 1}Department of Electrical and Electronic Engineering, The University of Hong Kong\\
      \textsuperscript{\rm 2}Department of Linguistics and Modern Languages, The Chinese University of Hong Kong\\

    \{motingyu, jcklam, vli\}@eee.hku.hk, yllcheung@cuhk.edu.hk
}

\begin{document}

\maketitle

\begin{abstract}
Alzheimer’s Disease (AD) is an irreversible neurodegenerative disease affecting 50 million people worldwide. Low-cost, accurate identification of key markers of AD is crucial for timely diagnosis and intervention. Language impairment is one of the earliest signs of cognitive decline, which can be used to discriminate AD patients from normal control (NC) individuals. Patient-interviewer dialogues may be used to detect such impairments, but they are often mixed with ambiguous, noisy, and irrelevant information, making the AD detection task difficult. Moreover, the limited availability of AD speech samples and variability in their speech styles pose significant challenges in developing robust speech-based AD detection models. To address these challenges, we propose DECT, a novel speech-based domain-specific approach leveraging large language models (LLMs) for fine-grained linguistic analysis and label-switched label-preserved (LSLP) data generation. 
Our study presents four novelties:  (1) We harness the summarizing capabilities of LLMs to identify and distill key Cognitive-Linguistic (CL) information (atoms) from noisy speech transcripts, effectively filtering irrelevant information. (2) We leverage the inherent linguistic knowledge of LLMs to extract linguistic markers from unstructured and heterogeneous audio transcripts. (3) We exploit the compositional ability of LLMs to generate LSLP AD speech transcripts consisting of diverse linguistic patterns to overcome the speech data scarcity challenge and enhance the robustness of AD detection models. (4) We use the augmented AD textual speech transcript dataset and a more fine-grained representation of AD textual speech transcript data to fine-tune the AD detection model. The results have shown that DECT, an integrated, LLM-assisted, speech-based AD detection model demonstrates superior model performance with an 11\% improvement in AD detection accuracy on the datasets from DementiaBank compared to the baselines.
\end{abstract}

\section{Introduction}
Alzheimer's disease (AD) is an irreversible neurodegenerative disorder affecting 50 million people worldwide. It progressively impairs cognitive functions, leading to memory loss, confusion, and difficulty in conducting daily activities normally. Early and accurate diagnosis of AD~\cite{nestor2004advances,venugopalan2021multimodal} is crucial for timely intervention and providing proper care to patients, potentially slowing the progression of the disease and improving their quality of life.\par
AD can be characterized by a range of biomarkers across different modalities~\cite{blennow2018biomarkers}, including demographic, cognitive, brain imaging, blood, cerebrospinal fluid (CSF), and genetic markers. 
Existing diagnostic methods for AD~\cite{young2024data} often rely on a combination of clinical assessments, neuropsychological tests, and brain imaging techniques. The most reliable biomarkers adopted by clinicians for AD detection are highly invasive, time-consuming, and expensive~\cite{laske2015innovative}. Brain imaging and CSF analysis methods face challenges when implemented on a large scale and may not effectively detect the early stages of disease progression. This underscores the need for more accessible and scalable diagnostic tools for detecting AD in the early phases of disease development~\cite{li2021designing}. Recent studies have highlighted language impairment as one of the earliest and obvious indicators of decline detectable among AD patients and the potential of using Cognitive-Linguistic (CL) markers for predicting AD onset and progression ~\cite{fraser2016linguistic,eyigoz2020linguistic}. These findings underscore the desirability of using speech or speech transcript for early AD detection and therapeutic treatment monitoring. Consequently, there is a growing interest in developing automated diagnostic tools to analyse linguistic patterns of AD speech and detect signs of AD onset. \par
One promising approach is to analyse the spontaneous speech data of AD patients~\cite{luz2021alzheimer,yang2022deep}. However, leveraging spontaneous speech or speech transcript for AD detection presents significant challenges. Existing spontaneous speech datasets, such as those from DementiaBank~\cite{becker1994natural}, are presented in dialogues, in a Q\&A format. They are limited in size and duration, lacking longitudinal temporal and demographic variation. The limited availability of labeled data and the variability in AD patients’ speech styles and linguistic patterns make it challenging to develop robust linguistic-based diagnostic models that can generalize well to diverse patient linguistic populations, which may require taking into account their population-specific linguistic features. Moreover, the speech transcripts of patient-interviewer dialogues often contain ambiguous, noisy, and irrelevant information, making it difficult to distinguish NC and AD individuals. The unstructured and heterogeneous natures of AD speech transcripts poses additional challenges for traditional NLP techniques to extract meaningful features for speech-based AD detection.\par
To address these challenges, we propose DECT, a novel domain-specific LLM-assisted approach that leverages the power of LLMs to distill linguistic and CL knowledge from spontaneous speech for robust speech-based AD detection. DECT presents a novel integrated LLM-assisted AD detection approach to \textbf{D}istill linguistic markers, \textbf{E}xtract CL Atoms, \textbf{C}ompose diverse speech transcript data, and ultimately \textbf{T}une task-specific models based on LLM-extracted linguistic markers and distilled CL Atoms. LLMs such as GPT models~\cite{ouyang2022training,achiam2023gpt},  have demonstrated remarkable success in performing various NLP tasks by capturing rich linguistic knowledge and context-dependent semantics, by utilizing the external and internal knowledge, together with in-context learning. Our work presents four novelties.
First, we leverage the inherent linguistic knowledge of LLMs to extract fine-grained linguistic marker from transcripts of AD speech to enable a more nuanced representation of AD speech transcript data. Second, we harness the summarizing capabilities of LLMs to identify and distill key CL information (atoms) from noisy transcripts, effectively filtering irrelevant information. Third, we exploit the compositional ability of LLMs to generate new AD speech transcript displaying diverse linguistic patterns based on the LLM-extracted linguistic markers and CL atoms, to overcome AD speech data scarcity and further enhance the robustness of task-specific model trained on limited speech transcript data. Fourth, we use the augmented speech transcript dataset and more fine-grained representation that blends linguistic markers with granular CL atoms for representing raw AD data to fine-tune the speech-based AD detection model. 

\section{Methodology}
\begin{figure*}[ht]
\centerline{\includegraphics[scale=0.55]{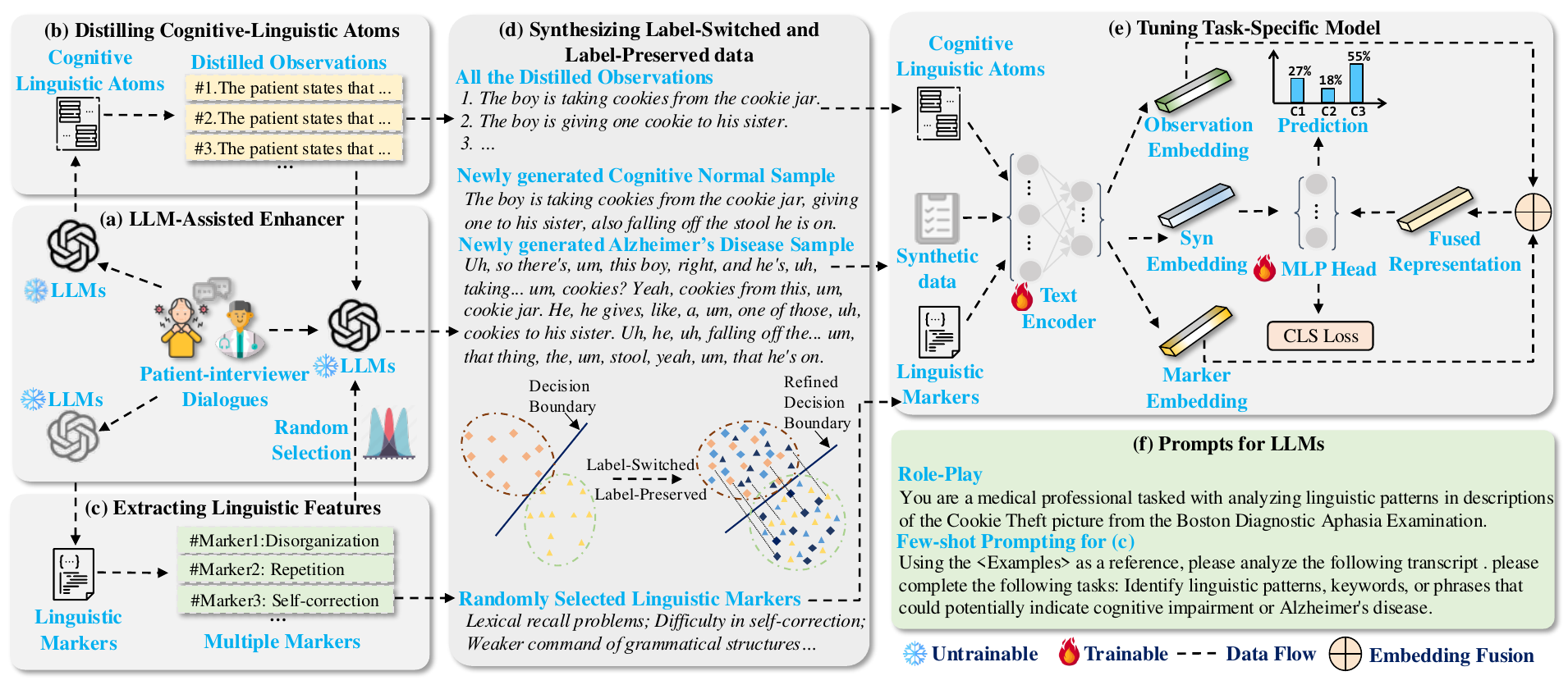}}
\caption{DECT: An LLM-assisted framework that leverages the inherent linguistic knowledge of LLMs to process spontaneous speech transcript. DECT presents a novel integrated LLM-assisted AD detection approach to \textbf{D}istill linguistic markers, \textbf{E}xtract Cognitive-Linguistic (CL) Atoms, \textbf{C}ompose diverse speech transcript data, and ultimately \textbf{T}une task-specific models based on LLM extracted linguistic markers and distilled CL atoms and generated data.}
\label{fig:frame_work}
\end{figure*}

As shown in Fig.\ref{fig:frame_work}, we propose DECT, a novel domain-specific approach leveraging LLMs for fine-grained linguistic analysis and LSLP data generation for AD detection. Our method employs LLMs to assist speech-based AD detection in a four-step process. First, distilling key CL information by filtering out irrelevant data. Second, extracting linguistic markers, fine-grained linguistic knowledge, from unstructured audio speech transcripts. Third, composing a nuanced representation by combining linguistic markers with CL atoms. Fourth, finetuning DECT, a task-specific LLM-assisted AD detection model using augmented and refined speech transcript data.
\subsection{Extracting Fine-grained Linguistic Knowledge from AD Speech Transcript}
A major challenge of speech-based AD detection being that the AD audio transcripts, in the form of Q\&A and dialogues are often ambiguous, noisy, and containing irrelevant information. Compared to LLMs~\cite{yang2023towards}, traditional NLP techniques often require extensive efforts to capture the subtle linguistic patterns and anomalies that may indicate cognitive impairment, particularly in the presence of noise. To address this challenge, we propose a novel integrated LLM-assisted approach that harnesses the inherent linguistic knowledge of LLMs to extract crucial linguistic markers, the fine-grained knowledge from audio AD speech transcripts. Few-shot prompting is a method that guides the LLM to perform specific tasks by keying into the GPT model a small number of instructions or prompts. In our case, we design prompts to instruct the LLMs to extract linguistic markers relevant to AD detection. We design the following prompt to guide LLMs:
\\
\textit{"Identify linguistic patterns, keywords, or phrases that could potentially indicate cognitive impairment or Alzheimer’s disease. "}\\
By providing these prompts, along with a few examples of audio speech transcripts labelled as AD or NC, we can guide the LLM to focus on the linguistic patterns that are most relevant for AD detection. The LLM then processes the speeches transcripts and generates a comprehensive set of linguistic markers of AD for each individual speech transcript. These linguistic markers include linguistic features such as: repetition, consistent flow of thought, sudden topic switching, etc.
\subsection{Identifying Cognitive-Linguistic Atoms via Distilling Raw AD Speech Transcript Data}
Distilled Cognitive-Linguistic (CL) atoms are concise, individual statements that capture key descriptive information from AD speech during a clinical interview. The concept is derived from "factual atoms", proposed by Stacey et al~\cite{stacey2023logical}, who used the term to represent the factual units derived from the premise of an utterance, in the field of Natural Language Inference (NLI). In this study, CL atoms serve as the base unit for objective analysis, distilling complex, noisy AD speech transcripts to provide clear, relevant pieces of CL information of AD patients. These atoms can be used to assess CL functions and detect early linguistic declines in AD patients. We leverage the advanced summarization and information extraction capabilities of LLM models to process raw audio AD speech transcripts. The LLMs are assigned with the tasks of identifying and distilling any salient CL information from the AD patients’ speech transcripts, derived from Cookie Theft, a CL assessment task. This involves filtering any irrelevant details, interviewer questions, and background noise from the speech dialogues. We design specific prompts to guide the LLM in extracting CL atoms from raw AD speech transcripts. The prompt is structured as follows:
\\
\textit{"Given the following transcript of a dialog between a patient and a doctor, extract the patient's speech into individual observations that capture the key information provided by the patient. Each fact should be a concise statement that preserves the relevant context from the conversation. \\Transcript: [example] \\The Patient's observations: [Generated cognitive-linguistic atoms]"}\\
This prompt instructs an LLM to focus on an AD patient’s raw speech, disregard the interviewer’s questions, and filter any noisy information to generate discrete, meaningful observations, which we defined as CL atoms. The LLM processes the speech transcript, identifying and removing irrelevant information, such as the interviewer’s questions and comments, filler words and phrases. The LLM then generates a list of CL atoms, each representing a single, coherent piece of information provided by the patient. These atoms are clear, standalone statements that capture the essence of the patient’s responses without requiring additional context from the transcript. By adopting this approach, we transform noisy, unstructured transcripts into a set of clear, relevant observations indicative of signs of AD onset.

\subsection{Generating Label-Switched and Label-Preserved Data}
Given the limited availability of labeled AD speech data and the high variability in AD speech patterns, models trained solely on incompletely sampled data may struggle to generalize across diverse patient populations. Synthesizing diverse in-distribution and unseen out-of-distribution data is crucial for enhancing model robustness in AD detection. By incorporating the newly generated data, we aim to improve the model's ability to handle unexpected speech patterns, rare linguistic features, and novel cognitive indicators that may not be well-represented in the original dataset. Thus, we utilize LLMs to generate new speech transcripts that simulate both AD and non-AD individuals. In our work, we generate new speech transcript by manually controlling the label perspectives, creating both label-switched and label-preserved (LSLP) data. This two-stage generation process involves (1) random decisions on switching (or not switching) labels, (2) incorporation of corresponding linguistic features. We designed prompts for LLMs to incorporate specific linguistic features of AD. Our prompt structure is as follows:\\
\textit{"Given a list of fact observations extracted from patients' speech transcript: [Distilled Cognitive-linguistic Atoms].\\ Based on the fact observations, generate a new speech transcript with the following linguistic features:[randomly selected linguistic markers]"}\\
The linguistic features extracted via LLM prompt can be divided into two categories. To incorporate AD linguistic markers, word-finding difficulties, semantic paraphrase, circumlocutions, reduced syntactic complexity, topic maintenance issues are included. For non-AD linguistic markers, detailed descriptions, complex sentence structures, rich vocabulary, coherent narrative have been included. To create a diverse AD speech transcript dataset, one can generate new speech data using inverted labels~\cite{peng2023generating}. To develop a new NC speech transcript dataset, we extract CL atoms from the original AD transcripts and inject non-AD features into these CL atoms and then reconstruct a new transcript consist of modified non-AD speech characteristics. Similarly, to develop a new AD speech transcript dataset, we also extract CL atoms from original NC transcripts and then randomly inject AD features into these CL atoms from an original NC speech transcript to generate new transcript with multiple combinations of AD speech characteristics, thus creating a rich variety of AD speech transcripts from one single NC speech transcript. This inverse operation helps the downstream model learn how to distinguish between genuine CL markers of AD from the superficial ones that may not be truly indicative of AD onset.

\begin{algorithm}[tb]
\caption{Optimization Process of DECT}
\label{alg:algorithm}
\textbf{Input}: Patient-interviewer dialogues $\mathcal{D} = \{d_1, ..., d_n\}; n \in (1,...,N)$\\
\textbf{Parameter}: $\bar{\theta}_{LLM}$ (frozen), $\theta_E$, $\theta_{MLP}$\\
\textbf{Output}: Prediction $\hat{y}$
\begin{algorithmic}[1]
\FORALL{$d_i \in \mathcal{D}$}
    \STATE $\mathbf{G}_{atom}^i, \mathbf{G}_{marker}^i \leftarrow f_{LLM}(d_i; \bar{\theta}_{LLM})$
    \STATE $\mathbf{E}_{atom} \leftarrow \mathbf{E}_{atom} \oplus f_E(\mathbf{G}_{atom}^i; \theta_E)$
    \STATE $\mathbf{E}_{marker} \leftarrow \mathbf{E}_{marker} \oplus f_E(\mathbf{G}_{marker}^i; \theta_E)$
    \STATE $\mathbf{S}_{syn}^i,i \leftarrow f_{LLM}(\mathbf{G}_{atom}^i, \mathbf{G}_{marker}^i; \bar{\theta}_{LLM})$
    \STATE $\mathbf{E}_{syn} \leftarrow \mathbf{E}_{syn} \oplus f_E(\mathbf{S}^i_{syn}; \theta_E)$
    \STATE $\mathbf{E}_{AMR} \leftarrow h_{fusion}(\mathbf{E}_{atom}, \mathbf{E}_{marker})$
    \STATE $\hat{y}, \mathcal{L}_{CLS}, \mathcal{L}_{syn} \leftarrow f_{MLP}(\mathbf{E}_{syn}, \mathbf{E}_{AMR}; \theta_{MLP})$
    \STATE $ \theta_E, \theta_{MLP} \leftarrow \text{argmin}_{\theta} [\mathcal{L}_{CLS}(\theta) + \mathcal{L}_{syn}(\theta)]$
\ENDFOR
\STATE \textbf{return} $\hat{y}$
\end{algorithmic}
\end{algorithm}
\subsection{DECT: An Integrated LLM-assisted AD Detection Approach}
As shown in Fig.\ref{fig:frame_work}, we present a novel AD detection approach for processing patient-interviewer dialogues, leveraging the power of LLMs. As illustrated in Algorithm \ref{alg:algorithm}, our approach aims to extract linguistic markers and CL atoms while generating diverse speech transcript for enhancing prediction robustness.
The framework begins with a set of patient-interviewer dialogues $\mathcal{D} = {d_1, ..., d_n}$, where $n \in (1,...,N)$. We employ a pre-trained LLM with frozen parameters $\bar{\theta}_{LLM}$ to distill two types of linguistic and CL cues from patient-interviewer dialogue, then map them into the CL atom embeddings $\mathbf{E}_{atom}$ and the linguistic marker embeddings $\mathbf{E}_{marker}$. These embeddings capture the factual CL information and linguistic features, respectively. Then, we fuse the atom embedding and the marker embedding as the representation of the raw patient-interviewer dialogues.\par
A key novelty of DECT is the incorporation of generated data with diverse linguistic features $\mathbf{S}_i$ based on all the extracted linguistic features by LLMs. The generated data with diverse linguistic patterns serves to enhance the model's generalization capabilities and the model's robustness to unforeseen speech patterns. DECT employs a text encoder with learnable parameters $\theta_E$ to further process the LLM-generated speech transcript information. The output of text encoder forms comprehensive representations  $\mathbf{E}_{atom}, \mathbf{E}_{marker}, \mathbf{E}_{syn} \in \mathbb{R}^{m \times d}$, where $m$ is the number of token and $d$ is the embedding dimension.
A fusion mechanism $h_{fusion}$ combines the CL atom and linguistic marker embeddings into a unified representation $\mathbf{E}_{fused}$, defined as Atom-Marker Representation (AMR). We concatenate the two embeddings and use a single dense layer to fuse the two. Subsequently, the output of this layer undergoes an additional GeLU activation function, resulting in $\mathbf{E}_{AMRep}$.\par
The final stage of DECT involves a multi-layer perceptron (MLP) with parameters $\theta_{MLP}$. This MLP takes both $\mathbf{E}_{AMRep}$ and $\mathbf{E}_{syn}$ as input, producing the prediction $\hat{y}$ along with two cross-entropy loss terms: $\mathcal{L}_{CLS}$ for AMR classification loss and $\mathcal{L}{syn}$ for generated data classification loss.
DECT parameters $\theta_E$ and $\theta_{MLP}$ are optimized jointly, minimizing a combined loss function. This multi-task optimization approach ensures that the AD detection model learns to process CL information while simultaneously adapting to potential distribution shifts in the input data.

\section{Experiments}
This part details how various experiments of DECT are conducted to enable linguistic-based AD detection. Our experiments are designed to assess the effectiveness of our LLM-enhanced approach across multiple dimensions, including model performance, component contributions and data generation strategies. We conduct our analysis on ADReSSo, a dominant dementia speech dataset for speech- and linguistic-based AD research. Our experimental framework encompasses three main aspects. Our DECT approach is compared with other baseline models. We evaluate different AD speech transcript data generation strategies and compare their performance and the impact of input representations. We conduct a systematic ablation study to understand the individual and combined effects of the key components of DECT. Through these experiments, we aim to demonstrate the robustness and efficacy of our method for AD detection.
\subsection{ADReSSo Dataset}
We use ADReSSo dataset~\cite{luz2021detecting}, a prominent AD audio speech dataset for AD research, derived from DementiaBank ~\cite{lanzi2023dementiabank} for our speech-based AD study. The ADReSSo dataset is available at https://dementia.talkbank.org. This dataset provides rich audio and complementary textual speech data, which can be further processed by LLMs to improve speech-based AD detection. The ADReSSo dataset serves as the benchmark for developing and evaluating AD detection methods through speech transcript analysis. ADReSSo focuses exclusively on spontaneous speech, making it a valuable resource to advance our understanding on what linguistic features are associated with cognitive declines among the AD patients. Participants are asked to perform a task using the Cookie Theft picture, as part of a CL assessment proposed by the Boston Diagnostic Aphasia Examination. During the exam, individuals are invited to describe what happens using a picture which depicts a complex scene, by giving a spontaneous speech that captures their individual linguistic abilities and cognitive functions. ADReSSo contains two groups of speech samples covering both Alzheimer’s Disease (AD) and Normal Control (NC) individuals. Our experimental subset includes 166 balanced samples: 86 ADs and 80 NCs. Each sample consists of a speech recording plus an equivalent speech transcript, providing a comprehensive basis for comparing the CL and the linguistic features between the ADs and the NCs.
\subsection{Experimental Settings}
Our experimental setup employs Python 3.8 on an Ubuntu 22.04 LTS operating system, utilizing PyTorch 1.9 and the Huggingface Transformers library. The experiments are performed on four NVIDIA GeForce RTX 4090 GPUs (24GB VRAM each). After performing experiments with alternative parameter values, we train the model using AdamW optimizer with a 15e-6 initial learning rate and 0.05 weight decay. The training process spans 50 epochs with a batch size of 8, employing a linear learning rate scheduler (warmup ratio 0.01)(see supplementary material for more details). For LLMs, we leverage a series of GPT model via the OpenAI API, generating responses with a fixed temperature of 1.
To ensure result reliability, we repeat all experiments 5 times using different random seeds, reporting the mean of performance and standard deviation of permanence metrics, including accuracy and F1-score. For baseline comparisons, we evaluate a series of BERT models alongside LLMs including Qwen~\cite{bai2023qwen}, ChatGLM3~\cite{team2024chatglm}, GLM4~\cite{team2024chatglm}, Yi-1.5~\cite{young2024yi}, GPT3.5~\cite{ouyang2022training}, GPT4~\cite{achiam2023gpt}, and GPT4o~\cite{achiam2023gpt}. To standardize the audios from the ADReSSo dataset, we use the Whisper model~\cite{radford2022whisper} for automatic speech recognition to transcribe the original audio recordings into text transcripts. In our primary results presented in Table \ref{tab:comparison with baseline methods} and Table \ref{tab:Ablation Study}, we employ a BioBERT model~\cite{lee2020biobert} as our text encoder for diagnostic task. 

\begin{table}[ht]
 \centering
\setlength{\tabcolsep}{1mm}
\scalebox{1}{
 \begin{tabular}{ccccc}\toprule
     \multirow{1}{*}{\textbf{Methods}}   & \multirow{1}{*}{\textbf{LLMs}}
     & \multicolumn{1}{c}{\textbf{Acc\%}} & \multicolumn{1}{c}{\textbf{F1\%}}\\ \midrule
        BERT-Base  & No &78.57\(\pm 2.38\) &74.28\(\pm 2.19\)\\
        BERT-Large  & No &80.95\(\pm 1.30\) & 76.30\(\pm 2.65\)\\
        FLAN-T5-Large  & No &75.23\(\pm 2.13\) &72.71\(\pm 3.18\)\\
        RoBERTa-Base  & No &83.33\(\pm 1.68\) & 81.61\(\pm 2.31\)\\
        RoBERTa-Large  & No &85.71\(\pm 2.38\) &84.09\(\pm 3.38\)\\ 
        BioBERT  & No &80.95\(\pm 2.71\) & 79.99\(\pm 2.27\)\\ 
        MedBERT  & No &84.28\(\pm 2.13\) & 82.37\(\pm 1.79\)\\ \hdashline

        Qwen1.5-7B-Chat  & Yes &46.39\(\pm 2.09\) &47.29\(\pm 2.83\)\\ 
        Qwen2-1.5B-Instruct  & Yes &50.75\(\pm 3.35\) &37.58\(\pm 7.97\)\\ 
        Qwen2-7B-Instruct  & Yes &51.45\(\pm 4.15\) &47.08\(\pm 5.39\)\\ 
        ChatGLM3-6B  & Yes &44.28\(\pm 4.24\) &41.15\(\pm 3.51\)\\
        GLM-4-9B-Chat  & Yes &40.97\(\pm 2.62\)&50.94\(\pm 5.12\)\\ 
        Yi-1.5-6B-Chat  & Yes &42.17\(\pm 0.85\) &44.82\(\pm 0.54\)\\
        Yi-1.5-9B-Chat-16K  & Yes &55.22\(\pm 0.92\) &59.21\(\pm 1.61\)\\
        GPT3.5  & Yes &58.11\(\pm 2.88\) & 47.32\(\pm 8.02\) \\ 
        GPT4  &  Yes &74.39\(\pm 1.50\) &77.69\(\pm 1.10\)\\ 
        GPT4o  & Yes &75.30\(\pm 1.10\) &78.31\(\pm 0.99\)\\ \hdashline
        DECT (GPT3.5) & Yes &83.33\(\pm 2.11\) & 80.85\(\pm 1.83\)\\
        DECT (GPT4) & Yes &85.31\(\pm 2.52\) & 82.46\(\pm 1.91\)\\ 
        DECT (GPT4o)  & Yes &\textbf{90.48}\(\pm \textbf{2.38}\) & \textbf{88.32}\(\pm \textbf{1.87}\)
    \\\bottomrule
 \end{tabular}
}
\caption{Performance of LLM- and Non-LLM Baselines vs DECT on ADReSSo Dataset.}
 \label{tab:comparison with baseline methods}
\end{table}
\subsection{Results and Analysis}
\subsubsection{Model Performance Comparison on AD Detection}
Table \ref{tab:comparison with baseline methods} presents the performance of DECT and other baselines on AD detection. The ADReSSo dataset is used for model training. To compare the performance of DECT for AD detection, different transformer models, including LLMs, have been used as the baselines (see Table 1). The results reveal several key insights into the effectiveness of different model architectures and the impacts of LLMs on AD detection tasks.
For non-LLM models, RoBERTa-Large demonstrates the best performance, achieving an impressive 85.71\% accuracy and 84.09\% F1 score. This is followed by MedBERT (84.28\%) and RoBERTa-Base (83.33\%), though MedBERT demonstrating a marginally superior F1 score. The strong performance of RoBERTa variants suggests that the pre-training approach is particularly effective for the task of AD detection.
There is a noticeable difference in performance among BERT variants. BERT-Large outperforms BERT-Base. As expected, domain-specific models such as BioBERT and MedBERT show mixed results. Interestingly, MedBERT performs significantly better than BioBERT, indicating that medical domain pre-training is more beneficial for AD detection than general biomedical pre-training.\par
We examine standalone performance of LLMs to understand their base capabilities in the AD detection task without additional fine-tuning or specialized architectures. The baseline helps us gauge the potential benefits of incorporating LLMs into more complex systems. We observe a significant drop in accuracy as compared to traditional transformer models. GPT4 and GPT4o show the best performance among LLMs, achieving similar accuracy (74.39\% and 75.30\%) and  F1 scores (77.69\% and 78.31\% respectively). Other LLMs including Qwen and ChatGLM3 variants perform notably worse, with accuracies ranging from 40.97\% to 55.22\%. This performance gap between standalone LLMs and traditional transformers suggests that while LLMs possess broader knowledge, they may struggle with specialized tasks without task-specific fine-tuning. \par
Our proposed method DECT leverages LLMs for data augmentation and BioBERT for textual data processing, showing competitive performance. The GPT4o variant of our approach achieves the highest accuracy (90.48\%) and F1 score (88.32\%) among all baseline methods. As we employ increasingly accurate and powerful LLMs, our results demonstrate a consistent performance enhancement progressing from GPT3.5 to GPT4 to GPT4o, demonstrating the benefits of more advanced LLMs in enhancing the text encoder's performance.
Notably, our best LLM-assisted variant (with GPT4o) outperforms the standalone BERT-base model by a significant margin (90.48\% vs 78.57\% accuracy), highlighting the effectiveness of our LLM-based LSLP strategy and AMR data representation. Our DECT method also outperforms most other models, including specialized ones with domain-knowledge such as MedBERT and BioBERT.\par
The findings above underscore the potential of LLM-assisted methods for low-cost accurate speech-based AD detection. The substantial performance improvement in the BERT encoder when combined with LLM-assisted techniques, suggests that an integrated approach can greatly enhance the accuracy of AD detection while preserving the computational efficiency of a smaller base model.

\begin{table}[ht]
 \centering
\scalebox{1}{
 \begin{tabular}{ccccc}\toprule
     \multirow{1}{*}{\textbf{Atom}} & \multirow{1}{*}{\textbf{Marker}} & \multirow{1}{*}{\textbf{LSLP}} 
     & \multicolumn{1}{c}{\textbf{Acc\%}} & \multicolumn{1}{c}{\textbf{F1\%}}\\ \midrule
        - & - & -  &80.95\(\pm 2.71\) & 79.99\(\pm 2.27\)\\ 
        $\surd$ & - & -  &82.38\(\pm 2.61\) & 84.95\(\pm 2.81\)\\ 
        - &$\surd$ & - &66.67\(\pm 2.38\) & 63.50\(\pm 4.86\)\\  
       -& - & $\surd$  &76.66\(\pm 2.61\) & 76.10\(\pm 2.51\)\\
       $\surd$ &$\surd$ & - & 84.12\(\pm 4.26\)& 85.65\(\pm 4.92\)\\ 
       $\surd$ &- & $ \surd$ &89.29\(\pm 1.31\) & 87.72\(\pm 0.78\)\\ 
       - &$\surd$ & $ \surd$ &76.19\(\pm 2.61\) & 72.61\(\pm 1.09\)\\ 
       $\surd$ &$\surd$ & $ \surd$ &\textbf{90.48}\(\pm \textbf{2.38}\) & \textbf{88.32}\(\pm \textbf{1.87}\)\\

    \bottomrule
 \end{tabular}
}
\caption{Ablation Study of Three DECT Components Based on ADReSSo Dataset (GPT4o).}
 \label{tab:Ablation Study}
\end{table}
\subsubsection{Ablation Study of DECT Components}

\begin{table}[ht]
 \centering

 \begin{tabular}{ccccc}\toprule
     \multirow{1}{*}{\textbf{LLMs}} & \multirow{1}{*}{\textbf{Input}} & \multirow{1}{*}{\textbf{Strategy}}
     & \multicolumn{1}{c}{\textbf{Acc\%}} & \multicolumn{1}{c}{\textbf{F1\%}}\\ \midrule
       GPT3.5 & Transcript & Mimic  &76.19 & 72.01\\ 
       GPT4 & Transcript & Mimic &75.00 & 71.36\\ 
       GPT4o & Transcript & Mimic  &83.33 & 81.34\\ 
       GPT3.5 & Transcript  & LSLP &84.12& 82.11\\  
       GPT4 & Transcript  & LSLP   &83.92 & 81.57\\ 
       GPT4o & Transcript  & LSLP  &84.12 & 83.13\\ 
       GPT3.5 & AMR & Mimic  &84.02 & 81.89\\
        GPT4 & AMR & Mimic &84.28 & 82.14\\ 
        GPT4o & AMR & Mimic  &88.88 & 86.06\\ 
       GPT3.5 & AMR & LSLP  &83.33 & 80.85\\ 
        GPT4 & AMR & LSLP &85.31 & 82.46\\ 
        GPT4o & AMR & LSLP  &\textbf{90.48} & \textbf{88.32} 
    \\\bottomrule
 \end{tabular}

\caption{Evaluation of MIMIC and LSLP Data Generation Strategies.}
 \label{tab:Synthetic Strategy}
\end{table}

Table~\ref{tab:Ablation Study} presents a comprehensive ablation study of DECT on the ADReSSo dataset, using GPT4o as the LLM. It examines the individual and combined effects of the three core components of DECT: CL Atom Extraction, Linguistic Marker Identification, and LSLP Data Generation. With reference to the ADReSSo dataset, the baseline model without adding any components achieves an accuracy of 80.95\% and an F1 score of 79.99\%. Data generation alone decreases the model performance of DECT to 76.66\% accuracy, suggesting that without properly injecting any linguistic features, the newly generated data may introduce noise. However, combining CL Atom and the LSLP data yields an accuracy of 89.29\%, a significant improvement over the baseline. Adding LSLP Data Generation to the two former strategies further boosts model performance to 90.48\% in accuracy and 88.32\% in F1 score, demonstrating the best model performance across all baselines. After conducting the ablation study, the following key observations have been noted: While individual components may occasionally decrease model performance, combining three LLM-based strategies consistently leads to an improvement in performance, highlighting the importance of synergistic interaction of different components, demonstrating the robustness of our DECT approach. The excellent performance of CL Atom Extraction and Linguistic Marker Identification suggests that proper representation is crucial for AD detection based on spontaneous speech data. While LSLP data generation plays a crucial role, its effectiveness is maximized when this strategy is used in parallel with fine-grained data representation.
\begin{figure}[htbp]
\centerline{\includegraphics[scale=0.47]{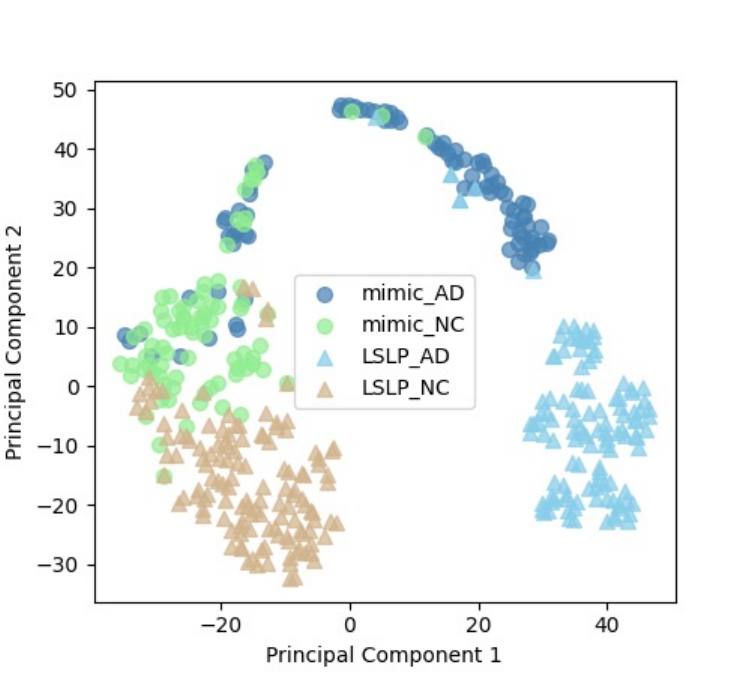}}
\caption{Visualization of Newly Generated AD and NC Speech Transcript Data with MIMIC Strategy and LSLP Strategy.}
\label{fig:synVSmimic}
\end{figure}
\subsubsection{Evaluation of Data Generation Strategies}
Table \ref{tab:Synthetic Strategy} presents a comprehensive evaluation of different data generation strategies applied to the ADReSSo dataset, revealing important insights into the effectiveness of input representation, and data generation strategies for AD detection. To compare the effectiveness of data generation strategy, we use the Mimic strategy as our baseline for comparison with our LSLP strategy. The Mimic strategy involves asking LLMs to replicate the style of the original transcript without altering attributes or labels. For input representation, AMR inputs generally yield better results compared to raw transcripts. For example, using GPT4o with the LSLP strategy, AMR inputs achieve 90.48\% accuracy compared to 84.12\% for transcript inputs. This indicates that the structured semantic information in AMR provides more valuable cues for AD detection. \par
Regarding the effectiveness of LSLP Data Generation strategy, the LSLP strategy generally outperforms the Mimic strategy, particularly for transcript inputs. The advantage is less pronounced for taking AMR as inputs, when both strategies perform comparably well in GPT3.5. This phenomenon may be caused by GPT3.5's inadequacy in generating high-quality AD speech transcripts. GPT4o consistently outperforms GPT4 and GPT3.5 across all configurations. This result suggests that more advanced LLMs contribute to better augmented data quality and improved AD detection performance. The highest accuracy (90.48\%) and F1 score (88.32\%) are achieved using GPT4o with AMR as inputs and the LSLP data generation strategy. The LSLP strategy generally outperforms the Mimic strategy, particularly for transcript inputs. This superior performance can be attributed to two key factors. First, the LSLP strategy generates synthetic samples that stay close to the decision boundary between the AD and the non-AD classes, helping the model learn more robust decision boundaries. Second, by switching and compositing the key linguistic attributes, the LSLP strategy can produce OOD samples that may not be found in the original dataset, enhancing the model's ability to generalize to unseen data patterns. These findings highlight the importance of the LSLP generation strategy when generating new AD and NC speech transcript data for AD detection.\par

Fig. \ref{fig:synVSmimic} presents a t-SNE visualization of synthetic data generated using mimic and LSLP strategies for AD and NC language patterns. The plot shows four distinct clusters: mimic AD, mimic NC, LSLP AD, and LSLP NC. LSLP (triangles) outperforms mimic (circles), with more compact and densely populated clusters for both AD (brown) and NC (light blue) labels. LSLP clusters occupy distinct regions in the lower left and right, while mimic clusters are more dispersed, with AD (blue) forming a curved pattern in the upper right and NC (light green) spread across the upper left. The more distinctive separation between AD and NC samples in LSLP suggests it  captures more effectively key differentiating characteristics of AD and NC language patterns. This visualization indicates that LSLP is superior in producing high-quality synthetic data that preserves the underlying structure of the original AD and NC language samples.\par 


\begin{figure}[h]
    
	\subfigure[Original Transcript]{
		\includegraphics[width=0.47\linewidth]{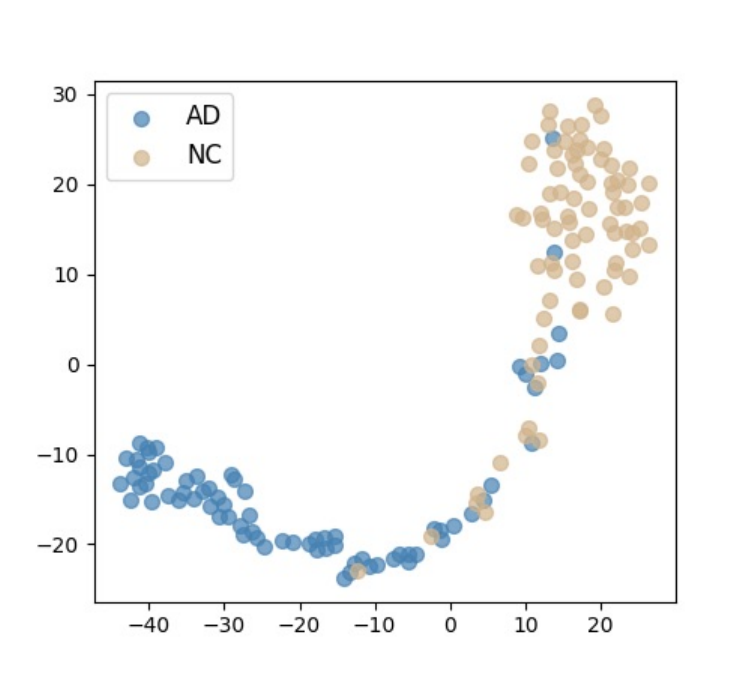}} 
	\subfigure[AM Representation]{
		\includegraphics[width=0.47\linewidth]{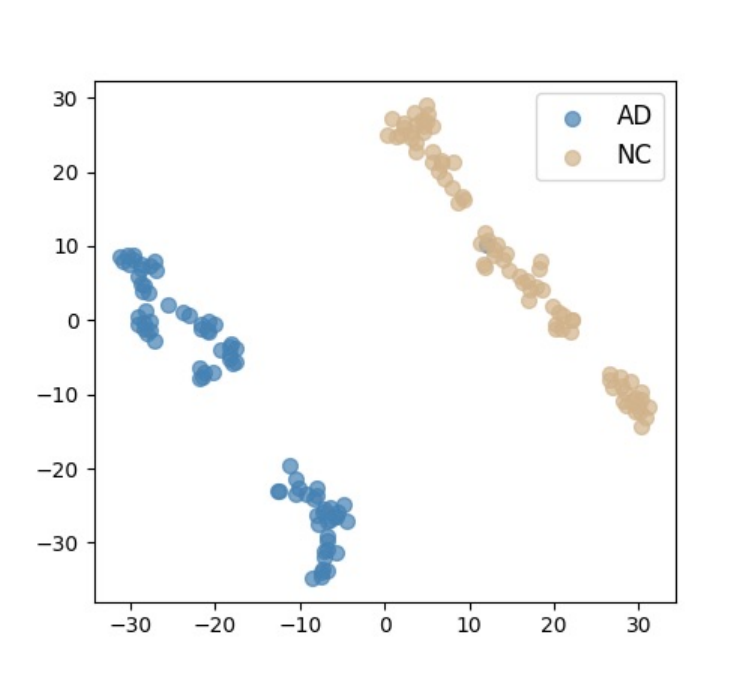}} 
    \caption{Embedding visualization: Comparison of AM Representation and Original Transcript using t-SNE for AD and CN groups}
    \label{fig:embedding_visualization}
\end{figure}

\subsubsection{Visualization of Embeddings}
To validate our hypothesis that leveraging LLM's knowledge can produce more fine-grained data representation for patient's spontaneous speech transcript, we use t-SNE to visualize the embeddings for AM Representation and the original transcript, as shown in Fig. \ref{fig:embedding_visualization}. The original 384-dimensional embeddings generated by the encoder have been reduced to two principal components for visualization. 
In the AM Representation plot (subfigure (b)), there is a distinctive separation between AD and NC clusters. The AD samples cluster mainly in the upper right quadrant, while the NC samples are distributed across the left hand side and the lower right hand side. This distinctive clustering suggests that AMR effectively captures differences between AD and NC subjects. Conversely, the original transcript plot (subfigure (a)) shows more overlaps between AD and NC groups, particularly in the central and the upper regions. While there is still some visible clustering, the separation is less pronounced when compared with AMR. This indicates that AMR may be more effective in discriminating ADs from NCs, when compared with the original transcript embeddings.
\begin{table}[htpb]
\centering

\renewcommand{\arraystretch}{0.8}
\scalebox{0.9}{%
\begin{tabular}{cp{0.85\columnwidth}}
\toprule
\textbf{No.} & \textbf{LLM-extracted CL Markers in AD} \\
\midrule
(1) & \textbf{Lexical recall problems:}
\begin{itemize}
    \item Hesitation, repetition, and frequent use of fillers
    \item Vague descriptions and erroneous word usage
    \item Difficulty in finding relevant words
\end{itemize} \\
\midrule
(2) & \textbf{Weak thought organization and thematic coherence:}
\begin{itemize}
    \item Unexpected topic switches
    \item Irrelevant past memories
    \item Mismatches between description and source image
\end{itemize} \\
\midrule
(3) & \textbf{Difficulty in self-correction:}
\begin{itemize}
    \item Struggle to correct errors or confusion
    \item May persist with incorrect statements
\end{itemize} \\
\midrule
(4) & \textbf{Weaker command of grammatical structures:}
\begin{itemize} 
    \item Inappropriate linguistic structures
    \item Difficulty in formulating grammatically correct sentences
\end{itemize} \\
\bottomrule
\end{tabular}%
}
\caption{Linguistic Features of AD Patients Identified by GPT4.}
\label{tab:linguistic_features}
\end{table}

\subsubsection{Linguistic Marker Analysis}
We summarized the crucial linguistic markers using GPT4 and identified the language features distinguishing AD from NC subjects. Table \ref{tab:linguistic_features} categorizes these extracted markers into four types, which align with existing literature \cite{glosser1991patterns,sajjadi2012abnormalities}. AD subjects exhibit: (1) lexical recall problems, leading to hesitation, repetition, and vague descriptions; (2) weak thought organization and thematic coherence, resulting in unexpected topic switches; (3) difficulties in self-correction; and (4) weaker command of grammatical structures. In contrast, NC subjects demonstrate fluent speech, coherent thought organization, ability to self-correct, and normal grammatical structures in their language production. We provide more detailed NC markers in the supplementary materials for a comprehensive comparison.

\section{Conclusion}
We propose DECT, an innovative approach for AD detection utilizing LLM-generated spontaneous speech data. We harness the capabilities of LLMs to extract CL atoms and linguistic markers from a balanced AD speech dataset ADReSSo. We capitalize on the capability of LLMs to generate new AD and NC speeches. To address the challenge of AD speech data scarcity, we developed an LSLP data generation strategy, generating label-switched and label-preserved (LSLP) AD and NC speech data samples, to augment the training data and improve model robustness. Experimental results have demonstrated a substantial improvement in accuracy, with an 11\% improvement for DECT over the baseline models. In the future, longitudinal analysis of AD linguistic markers can be made possible using LLM-assisted data generation techniques to generate more fine-grained speech transcript data over time. In addition, it is crucial to develop new methods that control the distributions of augmented/newly generated speech data to avoid task-specific models learning spurious correlation and reduce overfitting.

\section{Acknowledgments}
This work was supported in part by the United States National Academy of Medicine Healthy Longevity Catalyst Award (Grant No. HLCA/E-705/24), administered by the Research Grants Council of Hong Kong, and by The Hong Kong University Seed Funding for Collaborative Research 2023 (Grant No. 109000447).
\bibliography{aaai25}

\begin{thebibliography}{25}
\providecommand{\natexlab}[1]{#1}

\bibitem[{Achiam et~al.(2023)Achiam, Adler, Agarwal, Ahmad, Akkaya, Aleman,
  Almeida, Altenschmidt, Altman, Anadkat et~al.}]{achiam2023gpt}
Achiam, J.; Adler, S.; Agarwal, S.; Ahmad, L.; Akkaya, I.; Aleman, F.~L.;
  Almeida, D.; Altenschmidt, J.; Altman, S.; Anadkat, S.; et~al. 2023.
\newblock Gpt-4 technical report.
\newblock \emph{arXiv preprint arXiv:2303.08774}.

\bibitem[{Bai et~al.(2023)Bai, Bai, Chu, Cui, Dang, Deng, Fan, Ge, Han, Huang
  et~al.}]{bai2023qwen}
Bai, J.; Bai, S.; Chu, Y.; Cui, Z.; Dang, K.; Deng, X.; Fan, Y.; Ge, W.; Han,
  Y.; Huang, F.; et~al. 2023.
\newblock Qwen technical report.
\newblock \emph{arXiv preprint arXiv:2309.16609}.

\bibitem[{Becker et~al.(1994)Becker, Boiler, Lopez, Saxton, and
  McGonigle}]{becker1994natural}
Becker, J.~T.; Boiler, F.; Lopez, O.~L.; Saxton, J.; and McGonigle, K.~L. 1994.
\newblock The natural history of Alzheimer's disease: description of study
  cohort and accuracy of diagnosis.
\newblock \emph{Archives of neurology}, 51(6): 585--594.

\bibitem[{Blennow and Zetterberg(2018)}]{blennow2018biomarkers}
Blennow, K.; and Zetterberg, H. 2018.
\newblock Biomarkers for Alzheimer's disease: current status and prospects for
  the future.
\newblock \emph{Journal of internal medicine}, 284(6): 643--663.

\bibitem[{Eyigoz et~al.(2020)Eyigoz, Mathur, Santamaria, Cecchi, and
  Naylor}]{eyigoz2020linguistic}
Eyigoz, E.; Mathur, S.; Santamaria, M.; Cecchi, G.; and Naylor, M. 2020.
\newblock Linguistic markers predict onset of Alzheimer's disease.
\newblock \emph{EClinicalMedicine}, 28.

\bibitem[{Fraser, Meltzer, and Rudzicz(2016)}]{fraser2016linguistic}
Fraser, K.~C.; Meltzer, J.~A.; and Rudzicz, F. 2016.
\newblock Linguistic features identify Alzheimer’s disease in narrative
  speech.
\newblock \emph{Journal of Alzheimer's Disease}, 49(2): 407--422.

\bibitem[{Glosser and Deser(1991)}]{glosser1991patterns}
Glosser, G.; and Deser, T. 1991.
\newblock Patterns of discourse production among neurological patients with
  fluent language disorders.
\newblock \emph{Brain and language}, 40(1): 67--88.

\bibitem[{Lanzi et~al.(2023)Lanzi, Saylor, Fromm, Liu, MacWhinney, and
  Cohen}]{lanzi2023dementiabank}
Lanzi, A.~M.; Saylor, A.~K.; Fromm, D.; Liu, H.; MacWhinney, B.; and Cohen,
  M.~L. 2023.
\newblock DementiaBank: Theoretical rationale, protocol, and illustrative
  analyses.
\newblock \emph{American Journal of Speech-Language Pathology}, 32(2):
  426--438.

\bibitem[{Laske et~al.(2015)Laske, Sohrabi, Frost, L{\'o}pez-de Ipi{\~n}a,
  Garrard, Buscema, Dauwels, Soekadar, Mueller, Linnemann
  et~al.}]{laske2015innovative}
Laske, C.; Sohrabi, H.~R.; Frost, S.~M.; L{\'o}pez-de Ipi{\~n}a, K.; Garrard,
  P.; Buscema, M.; Dauwels, J.; Soekadar, S.~R.; Mueller, S.; Linnemann, C.;
  et~al. 2015.
\newblock Innovative diagnostic tools for early detection of Alzheimer's
  disease.
\newblock \emph{Alzheimer's \& Dementia}, 11(5): 561--578.

\bibitem[{Lee et~al.(2020)Lee, Yoon, Kim, Kim, Kim, So, and
  Kang}]{lee2020biobert}
Lee, J.; Yoon, W.; Kim, S.; Kim, D.; Kim, S.; So, C.~H.; and Kang, J. 2020.
\newblock BioBERT: a pre-trained biomedical language representation model for
  biomedical text mining.
\newblock \emph{Bioinformatics}, 36(4): 1234--1240.

\bibitem[{Li et~al.(2021)Li, Lam, Han, Cheung, Downey, Kaistha, and
  Gozes}]{li2021designing}
Li, V.~O.; Lam, J.~C.; Han, Y.; Cheung, L.~Y.; Downey, J.; Kaistha, T.; and
  Gozes, I. 2021.
\newblock Designing a protocol adopting an artificial intelligence (AI)--driven
  approach for early diagnosis of late-onset Alzheimer’s disease.
\newblock \emph{Journal of Molecular Neuroscience}, 71(7): 1329--1337.

\bibitem[{Luz et~al.(2021{\natexlab{a}})Luz, Haider, de~la Fuente, Fromm, and
  MacWhinney}]{luz2021detecting}
Luz, S.; Haider, F.; de~la Fuente, S.; Fromm, D.; and MacWhinney, B.
  2021{\natexlab{a}}.
\newblock Detecting cognitive decline using speech only: The adresso challenge.
\newblock \emph{arXiv preprint arXiv:2104.09356}.

\bibitem[{Luz et~al.(2021{\natexlab{b}})Luz, Haider, de~la Fuente~Garcia,
  Fromm, and MacWhinney}]{luz2021alzheimer}
Luz, S.; Haider, F.; de~la Fuente~Garcia, S.; Fromm, D.; and MacWhinney, B.
  2021{\natexlab{b}}.
\newblock Alzheimer's dementia recognition through spontaneous speech.

\bibitem[{Nestor, Scheltens, and Hodges(2004)}]{nestor2004advances}
Nestor, P.~J.; Scheltens, P.; and Hodges, J.~R. 2004.
\newblock Advances in the early detection of Alzheimer's disease.
\newblock \emph{Nature medicine}, 10(Suppl 7): S34--S41.

\bibitem[{Ouyang et~al.(2022)Ouyang, Wu, Jiang, Almeida, Wainwright, Mishkin,
  Zhang, Agarwal, Slama, Ray et~al.}]{ouyang2022training}
Ouyang, L.; Wu, J.; Jiang, X.; Almeida, D.; Wainwright, C.; Mishkin, P.; Zhang,
  C.; Agarwal, S.; Slama, K.; Ray, A.; et~al. 2022.
\newblock Training language models to follow instructions with human feedback.
\newblock \emph{Advances in neural information processing systems}, 35:
  27730--27744.

\bibitem[{Peng, Zhang, and Shang(2023)}]{peng2023generating}
Peng, L.; Zhang, Y.; and Shang, J. 2023.
\newblock Generating efficient training data via llm-based attribute
  manipulation.
\newblock \emph{arXiv preprint arXiv:2307.07099}.

\bibitem[{Radford et~al.(2022)Radford, Kim, Xu, Brockman, McLeavey, and
  Sutskever}]{radford2022whisper}
Radford, A.; Kim, J.~W.; Xu, T.; Brockman, G.; McLeavey, C.; and Sutskever, I.
  2022.
\newblock Robust Speech Recognition via Large-Scale Weak Supervision.

\bibitem[{Sajjadi et~al.(2012)Sajjadi, Patterson, Tomek, and
  Nestor}]{sajjadi2012abnormalities}
Sajjadi, S.~A.; Patterson, K.; Tomek, M.; and Nestor, P.~J. 2012.
\newblock Abnormalities of connected speech in semantic dementia vs Alzheimer's
  disease.
\newblock \emph{Aphasiology}, 26(6): 847--866.

\bibitem[{Stacey et~al.(2023)Stacey, Minervini, Dubossarsky, Camburu, and
  Rei}]{stacey2023logical}
Stacey, J.; Minervini, P.; Dubossarsky, H.; Camburu, O.-M.; and Rei, M. 2023.
\newblock Logical reasoning for natural language inference using generated
  facts as atoms.
\newblock \emph{arXiv preprint arXiv:2305.13214}.

\bibitem[{Team et~al.(2024)Team, Zeng, Xu, Wang, Zhang, Yin, Rojas, Feng, Zhao,
  Lai et~al.}]{team2024chatglm}
Team, G.; Zeng, A.; Xu, B.; Wang, B.; Zhang, C.; Yin, D.; Rojas, D.; Feng, G.;
  Zhao, H.; Lai, H.; et~al. 2024.
\newblock ChatGLM: A Family of Large Language Models from GLM-130B to GLM-4 All
  Tools.
\newblock \emph{arXiv e-prints}, arXiv--2406.

\bibitem[{Venugopalan et~al.(2021)Venugopalan, Tong, Hassanzadeh, and
  Wang}]{venugopalan2021multimodal}
Venugopalan, J.; Tong, L.; Hassanzadeh, H.~R.; and Wang, M.~D. 2021.
\newblock Multimodal deep learning models for early detection of Alzheimer’s
  disease stage.
\newblock \emph{Scientific reports}, 11(1): 3254.

\bibitem[{Yang et~al.(2023)Yang, Ji, Zhang, Xie, Kuang, and
  Ananiadou}]{yang2023towards}
Yang, K.; Ji, S.; Zhang, T.; Xie, Q.; Kuang, Z.; and Ananiadou, S. 2023.
\newblock Towards interpretable mental health analysis with large language
  models.
\newblock In \emph{The 2023 Conference on Empirical Methods in Natural Language
  Processing}.

\bibitem[{Yang et~al.(2022)Yang, Li, Ding, Xu, and Ling}]{yang2022deep}
Yang, Q.; Li, X.; Ding, X.; Xu, F.; and Ling, Z. 2022.
\newblock Deep learning-based speech analysis for Alzheimer’s disease
  detection: a literature review.
\newblock \emph{Alzheimer's Research \& Therapy}, 14(1): 186.

\bibitem[{Young et~al.(2024{\natexlab{a}})Young, Chen, Li, Huang, Zhang, Zhang,
  Li, Zhu, Chen, Chang et~al.}]{young2024yi}
Young, A.; Chen, B.; Li, C.; Huang, C.; Zhang, G.; Zhang, G.; Li, H.; Zhu, J.;
  Chen, J.; Chang, J.; et~al. 2024{\natexlab{a}}.
\newblock Yi: Open foundation models by 01. ai.
\newblock \emph{arXiv preprint arXiv:2403.04652}.

\bibitem[{Young et~al.(2024{\natexlab{b}})Young, Oxtoby, Garbarino, Fox,
  Barkhof, Schott, and Alexander}]{young2024data}
Young, A.~L.; Oxtoby, N.~P.; Garbarino, S.; Fox, N.~C.; Barkhof, F.; Schott,
  J.~M.; and Alexander, D.~C. 2024{\natexlab{b}}.
\newblock Data-driven modelling of neurodegenerative disease progression:
  thinking outside the black box.
\newblock \emph{Nature Reviews Neuroscience}, 25(2): 111--130.

\end{thebibliography}

\end{document}